\documentclass[letterpaper, 10 pt, conference]{ieeeconf}  % Comment this line out
\IEEEoverridecommandlockouts                              % This command is only
\usepackage{graphics} % for pdf, bitmapped graphics files
\usepackage{mathptmx} % assumes new font selection scheme installed
\usepackage{times} % assumes new font selection scheme installed
\usepackage{amsmath} % assumes amsmath package installed
\usepackage{amssymb}  % assumes amsmath package installed

\usepackage{url}
\usepackage{graphicx}
\usepackage{color}
\usepackage[squaren]{SIunits}

\newcommand{\refsec}[1]{Sect.~\ref{#1}}
\newcommand{\reffig}[1]{Fig.~\ref{#1}}
\newcommand{\reftab}[1]{Tab.~\ref{#1}}

\usepackage[noadjust]{cite}

\title{\LARGE \bf
Pose Estimation using Local Structure-Specific\\Shape and Appearance Context
}

\author{Anders Glent Buch$^{1}$, Dirk Kraft$^{1}$, Joni-Kristian Kamarainen$^{2}$, Henrik Gordon Petersen$^{1}$ and Norbert Kr\"{u}ger$^{1}$% <-this % stops a space
\thanks{$^{1}$Cognitive Vision Lab, University of Southern Denmark, 5230 Odense, Denmark. {\tt\small \{anbu,kraft,hgp,norbert\}$\texttt @$mmmi.sdu.dk}}%
\thanks{$^{2}$Tampere University of Technology, 33720 Tampere, Finland. {\tt\small joni.kamarainen$\texttt @$tut.fi}}%
}

\begin{document}

\maketitle
\thispagestyle{empty}
\pagestyle{empty}

%%%%%%%%%%%%%%%%%%%%%%%%%%%%%%%%%%%%%%%%%%%%%%%%%%%%%%%%%%%%%%%%%%%%%%%%%%%%%%%%

\begin{abstract}
We address the problem of estimating the alignment pose between two models using structure-specific local descriptors. Our descriptors are generated using a combination of 2D image data and 3D contextual shape data, resulting in a set of semi-local descriptors containing rich appearance and shape information for both edge and texture structures. This is achieved by defining feature space relations which describe the neighborhood of a descriptor. By quantitative evaluations, we show that our descriptors provide high discriminative power compared to state of the art approaches. In addition, we show how to utilize this for the estimation of the alignment pose between two point sets. We present experiments both in controlled and real-life scenarios to validate our approach.
\end{abstract}

%%%%%%%%%%%%%%%%%%%%%%%%%%%%%%%%%%%%%%%%%%%%%%%%%%%%%%%%%%%%%%%%%%%%%%%%%%%%%%%%

\section{Introduction}
The problem of determining the alignment transformation between two model surfaces has undergone extensive research in the computer vision community. In robotic manipulation applications, the instantiation of an object model into a scene is crucial to the success of further processing tasks. A relevant application is grasping and manipulation of objects, which requires a great deal of accuracy from the vision system. Accuracy in this process is also crucial for more high-level manipulation tasks, such as placing or assembly operations.

In the registration or stitching problem, multiple views of the same object or scene model are used for building a more complete representation. This requires very accurate alignments between the views in order for the result to be usable. The same method can be applied for the estimation of the external camera parameters in a multi-camera setup.

Methods for solving these problems have been applied both in the 2D image domain as well as in 3D using range data or RGB-D data where color information is also available. For these domains, a feature-driven approach is commonly taken, which is based on an attempt to remove false feature correspondences between the two models, thereby making it possible to compute the alignment transformation using the remaining true point to point correspondences. It is widely accepted that local descriptors provide the highest stability in this process due to their tolerance towards clutter and occlusions \cite{Osada2002}.

\begin{figure}[ht]
  \centering
  \includegraphics[width=\linewidth]{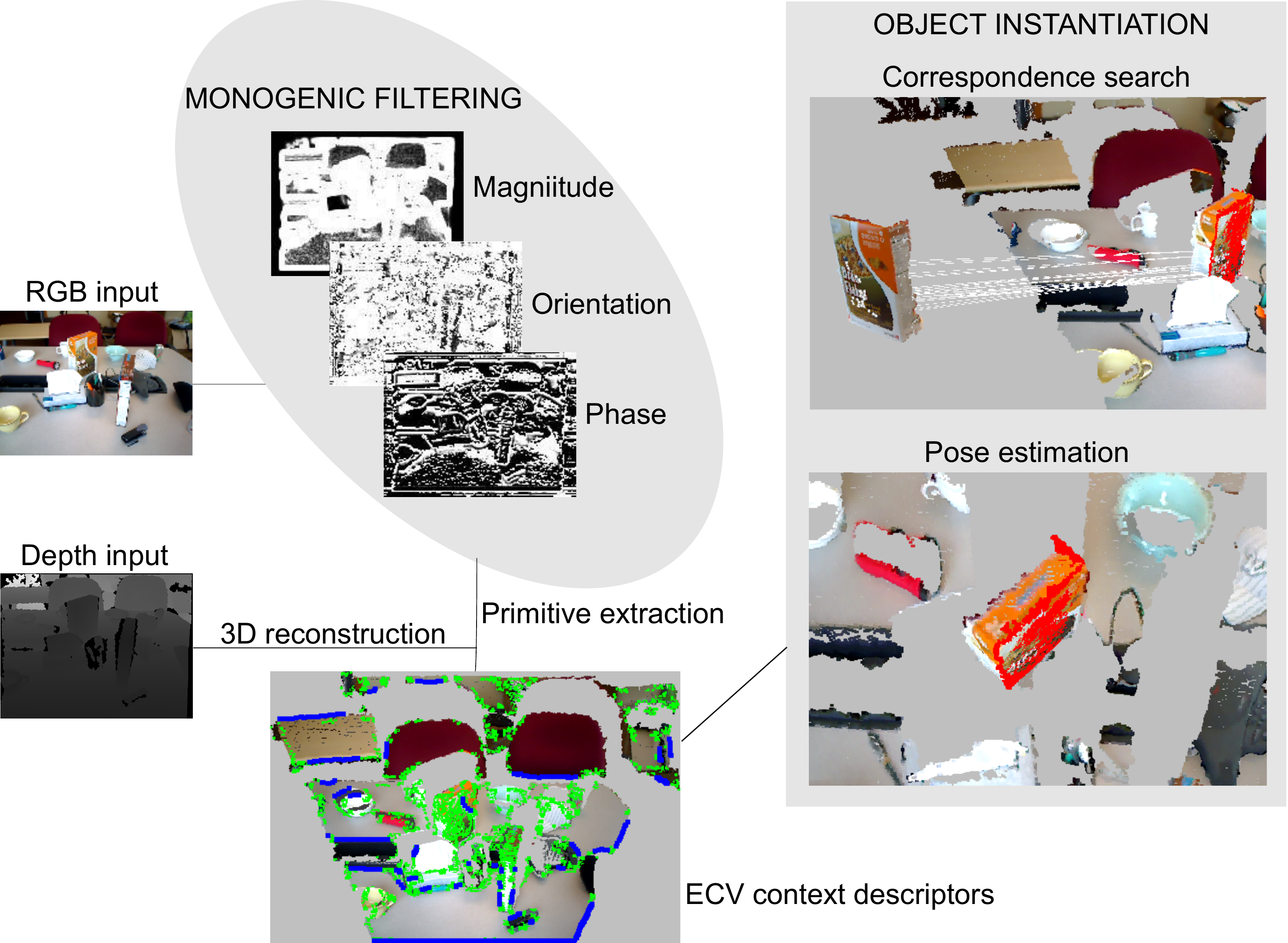}
  \caption{Processing pipeline of our pose estimation system. Left: input RGB-D images. Top middle: split of identity of appearance information by monogenic filtering, resulting in an image triple of local magnitude, orientation and phase. Bottom middle: ECV primitive extraction and their 3D reconstructed counterparts. Line segments are marked by blue, and texlets by green. Using this data, ECV context descriptors are generated. Top right: ECV context descriptor correspondences between a stored object model and the input scene. Bottom right: alignment pose estimation between the object and the scene.}
  \label{fig:pipeline}
\end{figure}

For image data, interest points or keypoints are often being detected and are afterwards augmented with contextual image information using the appearance of the local neighborhood for generating the final image descriptor \cite{Lowe1999,Belongie2002,MS05,Dalal2005,Bay2006surf}. The keypoint detection is done to locate distinct interest points, thus making the extraction process approximately viewpoint-invariant. Discriminative power is ensured by building a description of the appearance around the keypoint. Keypoint-based model descriptions tend to be quite sparse, since a higher density would decrease distinctiveness of the individual feature descriptor.

For 3D data or point clouds, a variety of shape descriptors have been developed over the last couple of decades \cite{Johnson1999,Hetzel2001,Frome2004,Rusu2009,Bariya2010,Drost2010}. These are often built for the complete data set, i.e. at every point on the model, although feature selection methods exist \cite{Gelfand2005,Rusu2008}. For shape descriptors, geometric invariants such as relative distances or angles can be used to describe the local neighborhood of a point. Since shape descriptors are often created for all points, these representations are dense.

In this paper, we introduce a new variant of descriptors for 3D models containing RGB data. The input to our feature processing can be either a dense stereo reconstruction or an RGB-D image (see \reffig{fig:pipeline}), providing both appearance and shape data. Our aim is to combine inputs from both the appearance and the shape domain in an efficient manner.

As will be detailed in \refsec{featuredescriptors}, keypoint detection is done by classifying individual pixels into different categories; this is achieved by the use of an Early Cognitive Vision (ECV) system. The density of an ECV representation falls between sparse image keypoint descriptors and dense 3D shape descriptors. This has the advantage that the model shape is captured in greater detail while maintaining a high level of discriminative power. To further increase the latter, we define context descriptors on top of the ECV model. As will be shown in the following sections, all this together allows for efficient pose estimation, both in terms of accuracy and speed.

Earlier works have applied ECV features for a range of computer vision applications. For the purpose of accurate scene representation, \cite{OlesenEtAl} presents a real-time extraction method for ECV texture primitives, providing uncertainty information for each primitive. In \cite{PilzEtAl09}, egomotion estimation is the target, and ECV edge structures are used for representing edge information in the observed scene. In \cite{DetryEtAl2009d}, ECV primitives are embedded in a probabilistic framework which allows for object recognition using generative models. Recently, ECV features have been applied for the task of grasping unknown objects \cite{Kootstra2012}, where a scene representation based on both edge and texture information is used for directly generating stable grasps.

The contribution of this paper is a model description containing both appearance and shape information and its applicability for different pose estimation tasks, such as object instantiation, scene registration and camera calibration as described in \refsec{experimentalresults}. We present an image processing pipeline which generates features both in the edge and in the texture domain. Using these features, we build a novel type of local 3D descriptor, utilizing the complementary power of both appearance and shape information. The result is a description with high discriminative power. We use our representation in a speed-optimized RANSAC \cite{FischlerBolles81} procedure, which shows the practical usability of our system. We argue for the efficiency of our representation in three ways: 1) we use both appearance and shape for describing a point, 2) keypoints are classified into edge/texture types, providing a structure-dependent descriptor, and 3) the keypoint density is high, allowing for more shape information than many other image descriptors.

We perform validation of our system both in a controlled experimental setup as well as for real-life scenarios. We compare against state of the art approaches and show that our representation can deal with a range of estimation problems more efficiently than other representations. We support this claim by showing 1) that our descriptors provide a high number of true correspondences under large viewpoint changes, and 2) how to efficiently solve the pose estimation problem between two models with large observation discrepancies.

The paper is structured as follows: \refsec{featuredescriptors} describes the acquisition of ECV context descriptors in a formal manner. In \refsec{poseestimation}, we motivate the use of our representation for efficient pose estimation. Experimental results are presented in \refsec{experimentalresults}, and conclusions are drawn in \refsec{conclusionsandfuturework}.

%%%%%%%%%%%%%%%%%%%%%%%%%%%%%%%%%%%%%%%%%%%%%%%%%%%%%%%%%%%%%%%%%%%%%%%%%%%%%%%%

\section{Feature descriptors}\label{featuredescriptors}
The work presented here builds on top of the observation space contained in an ECV model. We start by shortly describing the ECV feature extraction process and then move on to a presentation of our context descriptors.

\subsection{ECV primitives}
The atomic features created by the ECV process are referred to as \emph{primitives}. ECV primitives are extracted from an image by a rotation-invariant filtering method called the monogenic signal \cite{FelsbergSommer01} and later reconstructed in 3D. The monogenic signal filtering extracts a triplet of local magnitude, orientation and phase (MOP) at each pixel location, based on the spectrum of the appearance of the local neighborhood. This expansion of dimensionality is referred to as a \emph{split of identity}.

The MOP image is subdivided using a hexagonal grid, and keypoints are localized as the pixel with maximum magnitude response in each grid cell. The MOP splitting of an image motivates the concept of \emph{intrinsic dimensionality} (ID) \cite{FelsbergEtAl09}. The measure of spectral variance in a cell gives information about the dimensionality of the local image structure. A small variation in magnitude indicates a homogeneous patch, whereas a small variation in orientation indicates an edge or a line structure. Textured areas are characterized by large variations in both magnitude and orientation. By appropriately thresholding the ID values in a cell, the cell can be classified as belonging to either a homogeneous patch, an edge or a texture region \cite{PugeaultEtAl2010}. This gives low-level, but valuable information about what sensing modality different parts of the image represent. Since pixels belonging to homogeneous patches are inherently ambiguous, only edge regions and texture regions are used in this work.

\begin{figure}[ht]
  \centering
  \includegraphics[height=30mm]{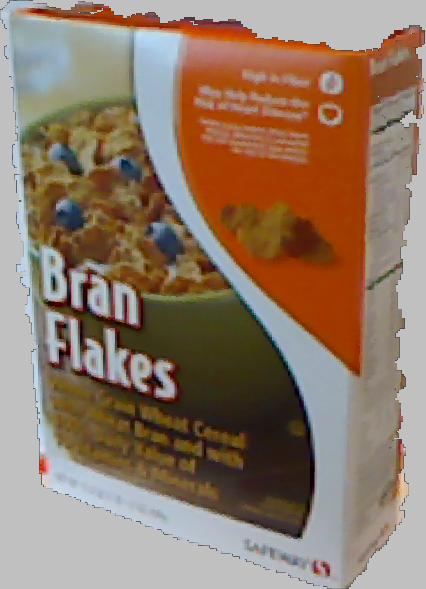}
  \quad
  \includegraphics[height=30mm]{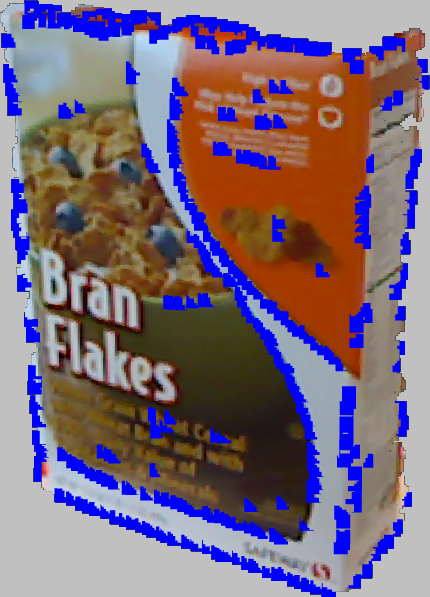}
  \quad
  \includegraphics[height=30mm]{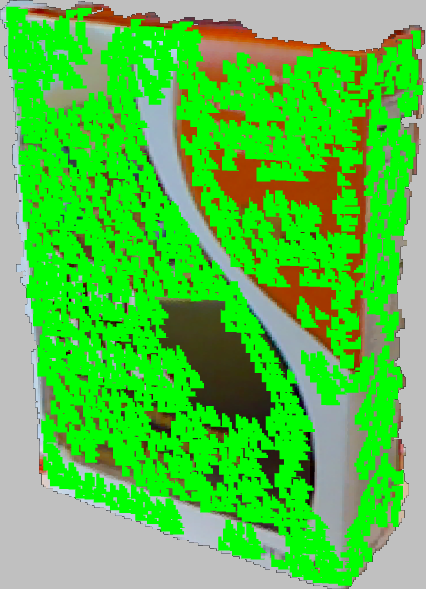}
  \caption{Visualization of different ECV features. Left: Input RGB-D model. Middle: ECV segments in blue. Right: ECV texlets in green.}
  \label{fig:cerealsegtex}
\end{figure}

\reffig{fig:cerealsegtex} shows a visualization of extracted edge/texture primitives of an example object. We refer to edge primitives as \emph{line segments}, or simply \emph{segments} (blue), and textured surface primitives as \emph{texlets} (green). In both this figure and in \reffig{fig:pipeline}, the 3D reconstruction process is done directly using the shape data from an aligned depth sensor. Recently, the extraction of 3D ECV primitives from RGB-D data has been implemented on GPU, allowing for real-time operation \cite{OlesenEtAl}.

Each ECV primitive includes a geometry and an appearance component. At this point, the geometry part consists of the 3D position and orientation. For segments, orientation is defined as the direction vector along the edge, and for texlets it is defined as the normal vector of the local surface region.

\subsection{ECV context descriptors}
\begin{figure}[ht]
  \centering
  \includegraphics[width=0.5\linewidth]{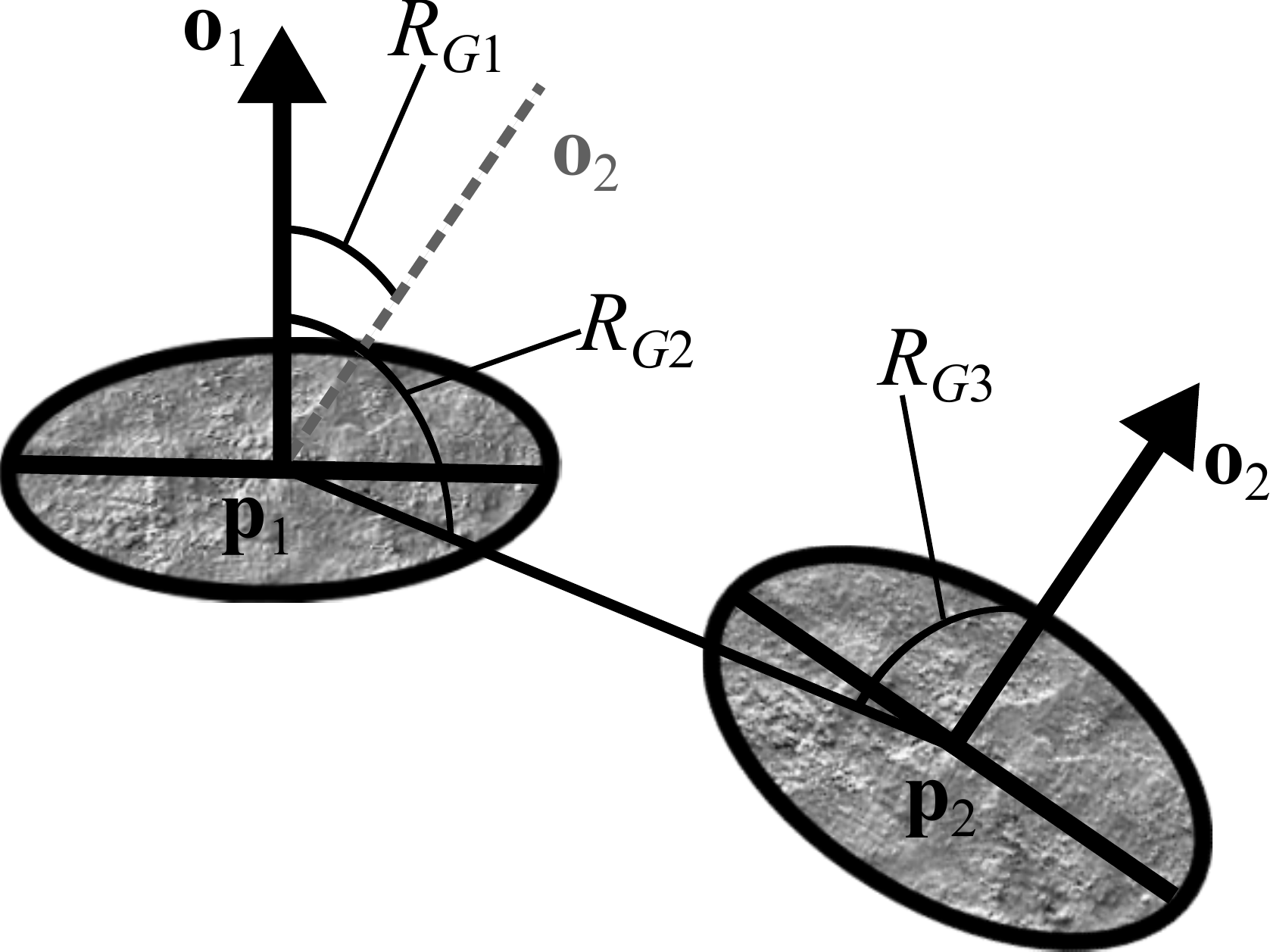}
  \caption{Three geometric relations used between feature points, in this case texlets with an associated normal vector. Note that we do not use the value of the angle, but instead the cosine for saving computation time.}
  \label{fig:geometricrelations}
\end{figure}

In order to increase the discriminative power of ECV features, we generate an appearance- and geometry-based description of the spatial neighborhood of a feature. At each feature location, we calculate relations between all feature point pairs on the ECV model inside a Euclidean neighborhood of radius $r$ and bin the results into histograms. We use the following geometric relations between two points $\mathbf{p}_1$ and $\mathbf{p}_2$ with orientations $\mathbf{o}_1$, $\mathbf{o}_2$:
\begin{itemize}
  \item Cosine between orientation vectors: $R_{G1} = \mathbf{o}_1 \cdot \mathbf{o}_2$.
  \item Cosine between first orientation vector and point to point direction vector: $R_{G2} = \mathbf{o}_1 \cdot \frac{\left( \mathbf{p}_2-\mathbf{p}_1 \right)}{\|\left( \mathbf{p}_2-\mathbf{p}_1 \right)\|}$.
  \item Cosine between second orientation vector and point to point direction vector: $R_{G3} = \mathbf{o}_2 \cdot \frac{\left( \mathbf{p}_2-\mathbf{p}_1 \right)}{\|\left( \mathbf{p}_2-\mathbf{p}_1 \right)\|}$.
\end{itemize}
Note that we use the general term \emph{orientation} vector, since this can be either a surface normal (texlets) or a direction (segments). In \reffig{fig:geometricrelations}, these geometric relations are visualized for a pair of texlets. An important detail is the ordering of any two points. The geometric relations $R_G$ are asymmetric in the sense that their sign changes depending on which feature point is regarded as the first $(\mathbf{p}_1)$ and the second $(\mathbf{p}_2)$. We resolve this by selecting $\mathbf{p}_1$ as the point closest to the source feature for which we are calculating the context descriptor.

For the appearance part, we create individual histograms for all three RGB color channels. As for the geometric relations, we take all possible point pairs in the region and calculate the three intensity gradients. For the three color channels, we denote these appearance relations as $R_{A1}$, $R_{A2}$ and $R_{A3}$. Again, we order points in pairs according to the distance to the source point. All in all, this gives six separate histograms, three geometry-based $R_G$ and three appearance-based $R_A$, which are all assembled in the final descriptor vector. Each histogram consists of 16 bins, resulting in an overall descriptor dimension of 96. We use $r = \unit{0.025}{\meter}$ throughout the rest of this paper, which we have found to be a reasonable compromise between locality and discriminative power.

We stress that in contrast with many other works on 3D shape description, our context descriptor formulation operates only on the ECV feature points (segments and texlets), and not on the underlying point cloud. This has two advantages: 1) the number of points in the neighborhood is reduced, leading to a computational speedup, and 2) by using points that are classified into line/texture structures, we avoid the use of homogeneous surface points, which do not add to the discriminative power. The latter statement is justified in \refsec{discriminativeanalysis}. We have deliberately kept the context descriptor formulation relatively simple, both to speed up the process, but also for showing the potential of ECV features for reliable context description. Clearly, this can be improved further by a systematic evaluation of alternative operators for describing feature contexts.

The extraction of a complete ECV context representation for a typical Kinect scene takes approximately \unit{1}{\second}, and we are currently working on porting the implementation to GPU for real-time extraction.

%%%%%%%%%%%%%%%%%%%%%%%%%%%%%%%%%%%%%%%%%%%%%%%%%%%%%%%%%%%%%%%%%%%%%%%%%%%%%%%%

\section{Pose estimation}\label{poseestimation}
In this section, we outline a simple, robust estimation algorithm for solving the pose estimation problem. Formally, the goal is to estimate a transformation $\hat{\mathbf{T}} \in SE(3)$ that minimizes the sum of squared distances between each point $\mathbf{p}$ on an object model $\mathbf{P}$ and the corresponding point $\mathbf{q}$ in the scene model $\mathbf{Q}$:
\begin{equation}
  \hat{\mathbf{T}} = \underset{\mathbf{T}}{\operatorname{argmin}} ~ \epsilon\left(\mathbf{T}\right) = 
               \underset{\mathbf{T}}{\operatorname{argmin}}\sum_{\mathbf{p}\in\mathbf{P}}\left(\mathbf{Tp-q}\right)^2
\end{equation}
In the above equation we use the homogeneous representation of points to allow for the matrix-vector multiplication. The problem stated above is often addressed using a robust and outlier-tolerant method, such as RANSAC. A common way of treating this issue is based on feature correspondences, where the following is run iteratively:
\begin{enumerate}
  \item Find $n \geq 3$ random object points in $\mathbf P$ and their corresponding points in $\mathbf Q$ by nearest neighbor matching of $SE(3)$-invariant feature descriptors. \label{item:hyp}
  \item Estimate a hypothesis transformation $\hat{\mathbf{T}}$ using the $n$ sampled correspondences. \label{item:est}
  \item Apply the hypothesis transformation to the object model $\mathbf{P}$. \label{item:apply}
  \item Find inlier points by spatial nearest neighbor search between the transformed object and the scene $\mathbf{Q}$, followed by Euclidean thresholding. If the number of inliers is too low, go back to step \ref{item:hyp}.
  \item Re-estimate a hypothesis transformation using the inlier point correspondences.
  \item Measure $\epsilon\left(\hat{\mathbf{T}}\right)$ using the inliers, and if it attains the smallest value so far, set $\hat{\mathbf{T}}$ as  the resulting transformation. \label{item:verify}
\end{enumerate}

In many cases, the algorithm can be optimized by stopping if $\epsilon$ falls below a predefined convergence threshold. Otherwise, the algorithm runs for the specified number of iterations. 

We apply a modification to the RANSAC pose estimation procedure above by enforcing a low-level geometric constraint after the first step of each iteration. To achieve this, we exploit the fact that distances are preserved under transformations by isometric elements of $SE(3)$. Specifically, we make a simple check of the ratio between the edge lengths of the virtual polygons formed by the $n$ sampled points on both the object and the scene models. We denote the points sampled by feature correspondences as $\mathbf{p}_i,\mathbf{q}_i, ~ i \in \lbrace 1,\ldots,n \rbrace$. The edge lengths on the object polygon are given as $d_{\mathbf{p},i}=\|\mathbf{p}_{i+1\bmod n}-\mathbf{p}_i\|$, likewise for the scene polygon edge lengths $d_{\mathbf{q},i}$. We then calculate the relative dissimilarity vector $\delta$ by ratios between the $n$ polygon edge lengths:
\begin{equation}
  %\delta = \begin{bmatrix} \left| 1-\frac{d_{\mathbf{q},1}}{d_{\mathbf{p},1}} \right| & \ldots & \left| 1-\frac{d_{\mathbf{q},n}}{d_{\mathbf{p},n}} \right| \end{bmatrix}
  \delta = \begin{bmatrix} \frac{\left|d_{\mathbf{p},1}-d_{\mathbf{q},1}\right|}{\max\left(d_{\mathbf{p},1},d_{\mathbf{q},1}\right)} & \ldots & \frac{\left|d_{\mathbf{p},n}-d_{\mathbf{q},n}\right|}{\max\left(d_{\mathbf{p},n},d_{\mathbf{q},n}\right)} \end{bmatrix}
\end{equation}
In case of a perfect match between two polygons, $\delta$ is identically zero. In practice, we can expect the largest deviation to be below a certain threshold $t_{poly}$:
\begin{equation}
  \| \delta \|_{\infty} \leq t_{poly}
\end{equation}
Our modification is therefore to insert the following step between step \ref{item:hyp} and \ref{item:est}:
\begin{itemize}
  \item Calculate the dissimilarity vector $\delta$ between the sampled polygon edge lengths. If $\| \delta \|_{\infty} > t_{poly}$, go back to step \ref{item:hyp}.
\end{itemize}
This verification step is significantly cheaper than enforcing the full geometric constraint using steps \ref{item:est}-\ref{item:verify}. Under the assumption that this step does not filter out hypothesis poses that align the models correctly, we can expect the same probability of success, but in much shorter time. Clearly, this assumption only holds as long as we can expect fairly accurate geometric observations such that the polygons are indeed isometric. If the sensor used exhibits a large depth error or produces a large distortion effect, the threshold $t_{poly}$ would need to be set higher to allow for these inaccuracies. With the quality of sensors available today, we expect this problem to be of limited extent. We use $t_{poly}=0.25$, thereby allowing for a maximal edge length dissimilarity of \unit{25}{\%}.

A large array of modifications to the original RANSAC has been proposed. These are based on different strategies, e.g. preemption \cite{nister2005preemptive}, local optimization \cite{Chum2003} and progressive sampling \cite{Chum2005}, only to name a few. Our modification is based on the simple criterion that wrong hypotheses should be filtered out immediately, thus saving time for generating a higher number of hypothesis poses. Our approach can be regarded as being hierarchical in the rejection phase in the sense that we introduce a preliminary low-level polygon-based rejection, which does not require a pose, before the usual inlier-based rejection phase.

The formula for calculating the required number of RANSAC iterations $k$ given a desired success probability $p$ and an expected inlier fraction $w$ is \cite{FischlerBolles81}:
\begin{equation}
  k = \frac{\log\left(1-p\right)}{\log\left(1-w^n\right)}
\end{equation}
In our experiments, we sample $n=3$ points in step \ref{item:hyp}. We use a conservative inlier fraction estimate of $w=0.05$ and a desired success rate of $p=0.99$, giving $k \approx 37000$. We set the Euclidean inlier threshold to \unit{0.01}{\meter}, and the required number of inlier points is set to \unit{50}{\%} of the total number of object model points.

%%%%%%%%%%%%%%%%%%%%%%%%%%%%%%%%%%%%%%%%%%%%%%%%%%%%%%%%%%%%%%%%%%%%%%%%%%%%%%%%

\section{Experimental results}\label{experimentalresults}
We present three different experimental validations of our approach. In \refsec{discriminativeanalysis}, we present a systematic evaluation method for determining the amount of true feature correspondences between two models. In \refsec{srac}, we apply our estimation procedure to the problem of registering two views of a scene. Finally, we show in \refsec{realexperiments} results from our own setup. All experiments have been performed on a PC equipped with an \unit{2.2}{\giga\hertz} Intel i7 processor with four cores (i7-2720QM). For speeding up nearest neighbor searches, we use the Fast Library for Approximate Nearest Neighbors \cite{muja_flann_2009}.

\subsection{Discriminative analysis}\label{discriminativeanalysis}
A key element of efficient pose estimation is the establishment of correspondences between the models to be aligned. In this initial experiment, we outline an algorithm for testing correspondences between two models. In general, we cannot a priori determine whether two points correspond; indeed, this is the goal of the estimation process. In \cite{Lowe04}, a correspondence is established during estimation if the ratio between the closest and the second closest feature matching distance is low. This ensures uniqueness of a correspondence, but only increases the probability of a correspondence actually being true.

\begin{figure*}[ht]
  \includegraphics[width=\linewidth]{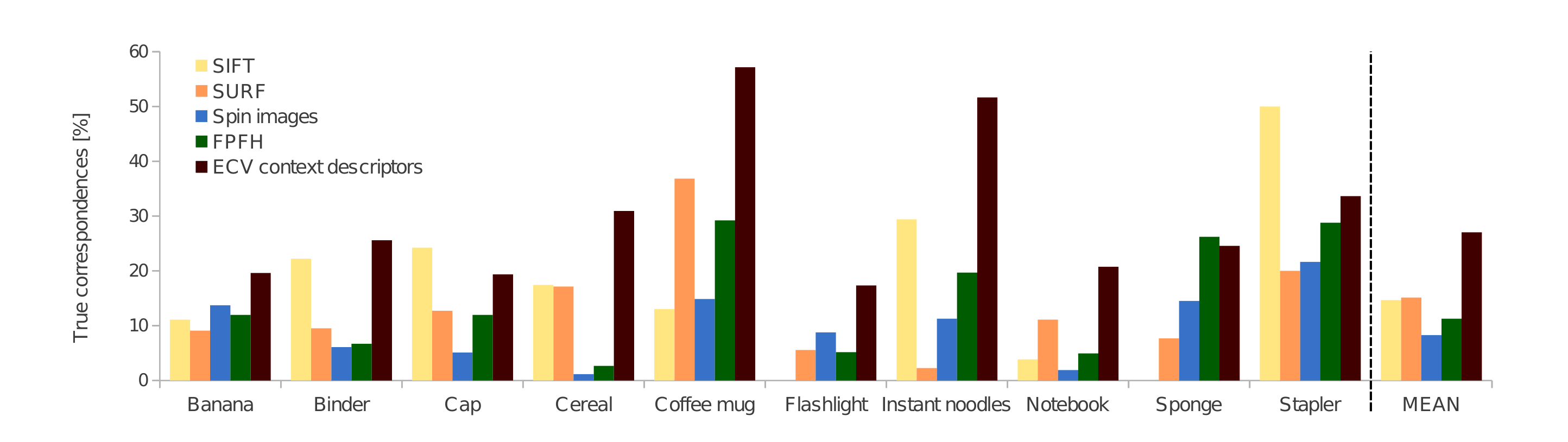}
  
  \hspace{60pt}\includegraphics[width=0.73\linewidth]{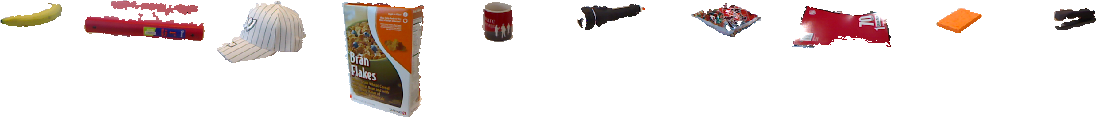}
  
  \caption{Fraction of true correspondences using five different descriptors, including our own, for ten exemplary objects. The rightmost part of the chart shows the mean over all test cases.}
  \label{fig:correspondences}
\end{figure*}

Here we reverse the process: we start by performing alignment of the two models, and afterwards we check which of the originally calculated feature correspondences were correct, simply by thresholding the Euclidean distance between point pairs matched in the initial correspondence calculation phase. This is inspired by the correspondence rejection implemented in the Point Cloud Library (PCL) \cite{Rusu_ICRA2011_PCL}. More specifically, we do the following:
\begin{enumerate}
  \item Generate feature descriptors for both models and calculate the nearest matching scene feature of each object feature. This is the initial, hypothesized correspondence set $C_{hyp}$. \label{item:corr}
  \item Perform accurate alignment of the two models such that each object surface point is well aligned to its true corresponding scene surface point (see text below). \label{item:align}
  \item Loop over all hypothesized correspondences and perform thresholding: if the Euclidean distance between the aligned matched point pair is below a given threshold, store this as a true correspondence. The result is a reduced set of correspondences $C_{true}$, likely to be true correspondences. \label{item:threscorr}
  \item Calculate the \emph{true correspondence} score as $C_{true}/C_{hyp}$.
\end{enumerate}
If the models are sufficiently close to begin with, a local method such as Iterative Closest Point (ICP) \cite{Besl1992,Zhang1994} can be used for step \ref{item:align}. Otherwise, a robust estimation method such as RANSAC can be applied, based on the correspondences generated in step \ref{item:corr}. The Euclidean threshold in step \ref{item:threscorr} should be set to at least the expected mean accuracy of the alignment in order to compensate for estimation inaccuracies.

\begin{figure}[ht]
  \centering
  
  \includegraphics[width=\linewidth]{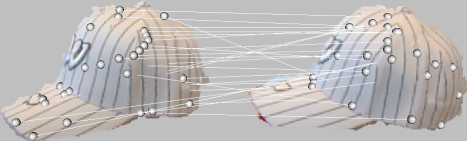}
  
  \vspace{5pt}
  
  \includegraphics[width=\linewidth]{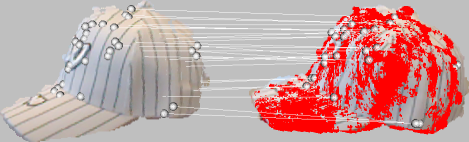}
  
  \caption{Top: ECV context descriptor correspondences between the two object models. Bottom: filtered correspondences between the two object models after ICP alignment (aligned leftmost model overlaid in red) and Euclidean distance thresholding. For displaying purposes, only 25 randomly selected correspondences are shown in both cases.}
  \label{fig:capcorr}
\end{figure}

In \reffig{fig:capcorr}, we show the whole process of alignment-based correspondence rejection between two views of the same object. The top figure shows a subset of the initial hypothesis set $C_{hyp}$ by white lines, and the bottom figure a subset of the remaining correspondences $C_{true}$, again by white lines, as well as the ICP result by overlaying the aligned model in red.

The objects in both this figure and in Figs.~\ref{fig:pipeline} and \ref{fig:cerealsegtex} are taken from the RGB-D database \cite{Lai2011} containing Kinect views of a large number of object models as well as some example scenes. All objects in the database are captured on a turntable with only a few degrees displacement between frames, and from three different elevations, giving several hundred views per object. Here we used the 1. and the 15. frame, which are displaced by approximately \unit{30}{\degree}.

We have carried out this procedure for a set of randomly chosen objects from the RGB-D database, all using two views separated by 15 frames. For each object, we calculate the correspondence score using a Euclidean threshold of \unit{0.01}{\meter}. We compare our ECV feature descriptors against state of the art descriptors in the image domain (SIFT \cite{Lowe1999} and SURF \cite{Bay2006surf}) as well as in the 3D shape domain (Spin images \cite{Johnson1999} and FPFH \cite{Rusu2009}). For SIFT/SURF, we use OpenCV with standard settings. SURF does require a user-specified Hessian threshold for the keypoint detection, which we set to 500. For the shape descriptors, we use the PCL implementations with the radius set to $r=\unit{0.025}{\meter}$. The calculated correspondence scores are reported in \reffig{fig:correspondences}, with the mean over all scores shown in the rightmost part.

The results in \reffig{fig:correspondences} are not surprising. Indeed, shape-dominant objects such as ``Banana'' and ``Sponge'' are best described using shape descriptors. On the other hand, texture-rich objects such as ``Binder'' and ``Cereal'' are best captured by the appearance-based image descriptors. In one instance, ``Sponge'', the FPFH shape descriptor provided slightly better discriminative abilities than our proposed descriptor. This is expected since the appearance part of this object is highly ambiguous. For the objects ``Cap'' and ``Stapler'', SIFT produced better results. In the latter case, however, SIFT descriptors are calculated at only eight detected keypoints, making the object model extremely sparse. In comparison, ECV processing produces 113 keypoints for this object. We conclude from this study that our descriptors outperform the others in general, meaning on average over a range of objects with different shapes and appearances. The mean correspondence score from the total set is \unit{27}{\%}, whereas the next best, SURF, has a mean score of \unit{15}{\%}.

\subsection{Scene registration and calibration}\label{srac}
Algorithms for registering or stitching multiple images of a scene have been successfully implemented using robust 2D descriptors \cite{shum2000systems,brown2003recognising}. The registration of multiple 3D models also has its practical usage, e.g. for building a model from several views. Another application is the calibration of a multi-camera setup, where multiple views of a scene must be registered in order to estimate the relative camera poses.

\begin{figure}[ht]
  \centering
  \includegraphics[width=\linewidth]{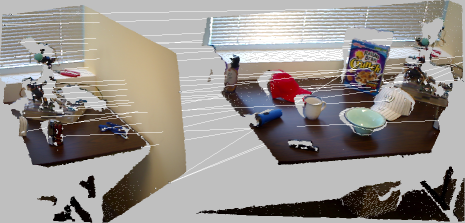}
  
  \vspace{5pt}
  
  \includegraphics[width=0.4925\linewidth]{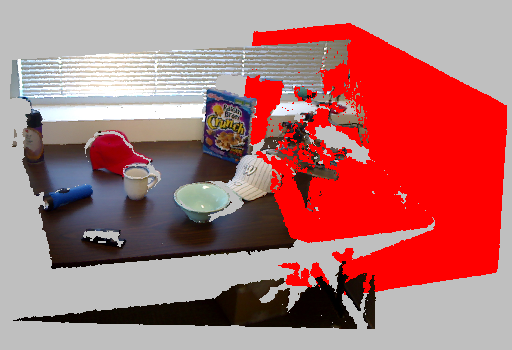}
  \includegraphics[width=0.4925\linewidth]{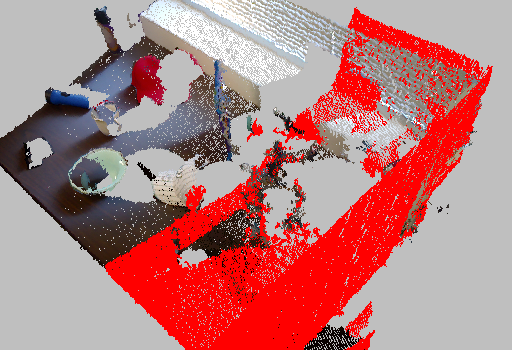}
  
  \caption{Registration of two different views of the same scene using our optimized RANSAC algorithm. Top: input Kinect views, showing 25 randomly selected correspondences computed using ECV context descriptors. Bottom left: registration result with the aligned version of the leftmost scene view in the top image overlaid in red. Bottom right: zoom of the aligned point sets. Note the high quality of the alignment, which is performed without refinement.}
  \label{fig:scenes}
\end{figure}

If enough shape information is available in the scene, the relative camera transformation can be determined very accurately without the use of markers. In \reffig{fig:scenes}, we show the result of applying our modified RANSAC to two views of a scene with a wide baseline separation. For these kind of tasks, we lower the required number of inliers due to the limited overlap. In this experiment, we set the inlier fraction to \unit{10}{\%} of the number of points in the leftmost scene.

We stress that we do not perform refinement of the results reported here, yet due to the high number of iterations, we achieve very accurate alignments. The registration time is below \unit{1}{\second}, which is achieved partly by our modified RANSAC procedure, partly by the low density of the ECV representation. The leftmost and rightmost scene in \reffig{fig:scenes} are described by 1255 and 1448 ECV context descriptors, respectively.

In \reftab{tab:scenes}, we show two relevant statistics of our modified RANSAC compared to the standard RANSAC algorithm. The numbers are the means of 100 runs of the above estimation problem. To be able to compare timings, we set the iteration count to 5000 for both cases. We observe that on average, our RANSAC performs almost 15 times faster than standard RANSAC. This is achieved without compromising the quality of the alignment, as can be seen by the mean fit error, which is calculated as the mean $\epsilon$ over the number of thresholded inliers.

\begin{table}[h]
  \centering
  \caption{Timings and mean inlier fit errors of our modified RANSAC compared to standard RANSAC, computed using 100 runs of the scene registration problem in \reffig{fig:scenes}.}
  \label{tab:scenes}
  \begin{tabular}{|c|c|c|}
    \hline
                    & Run time [s] & Mean fit [m] \\
    \hline
    RANSAC          & $13.68$ & $6.40 \cdot 10^{-3}$ \\
    Modified RANSAC & $0.88$ & $6.31 \cdot 10^{-3}$ \\
    \hline
  \end{tabular}
\end{table}

\subsection{Experiments from real setup}\label{realexperiments}
In this section, we present object pose estimation results from a robotic setup in our own laboratory. For all the experiments, we have conducted, estimation time for a single object has remained below \unit{0.1}{\second}. We currently use the pose estimation algorithm presented in this work for benchmarking robotic grasping strategies developed at our institute

\subsubsection{Model representation} The object models given are taken from the KIT database of full textured CAD models generated using a triangulation scanner and a high resolution camera \cite{Kasper2012}. We have acquired physical copies of a subset of the objects, which we use for testing. Since ECV processing requires images for capturing the appearance, we render the objects from four different views. The extracted ECV features from a view are backprojected to the 3D model shape, after which the context descriptors are built for that view. During pose estimation, we use the view with the best match in the scene, if such a match exists.

\subsubsection{Color calibration} An important practical consideration for this experiment is the fact that there may be discrepancies between the color representations in the provided object models and the scenes captured in our setup. This is due to the fact that the object/scene models are captured under different lighting conditions. Since we are not interested in attempting to copy the illumination conditions used when capturing the objects, we devised a simple solution for aligning the color spaces.

We assume that an RGB color triplet $\mathbf{c}_{KIT}$ from a given KIT object model has undergone a linear mapping $\mathbf{A}$ up to the point where it is observed as $\mathbf{c}_{SDU}$ by our sensor:
\begin{equation}
  \mathbf{Ac}_{KIT} = \mathbf{c}_{SDU}
\end{equation}
By estimating $\mathbf{A}$, which is a 3-by-3 matrix, we can transform the color values of the stored object models such that they align with our color conditions. We estimate $\mathbf{A}$ once, simply by labeling a set of corresponding pixel RGB values between an object model and a scene from our laboratory containing that object. $\mathbf{A}$ has nine free parameters, and we have used ten color pixels for estimating it. The resulting color calibration matrix gives the required color space alignment for obtaining valid correspondences using ECV context descriptors.

\begin{figure}[ht]
  \centering
  \includegraphics[width=0.95\linewidth]{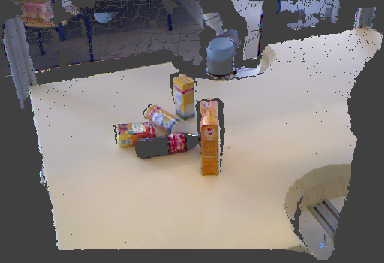}
  
  \vspace{5pt}
  
  \includegraphics[width=0.2\linewidth]{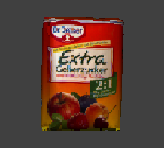}
  \hspace{5pt}
  \includegraphics[width=0.2\linewidth]{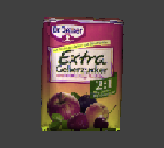}
  \hspace{5pt}
  \includegraphics[width=0.2\linewidth]{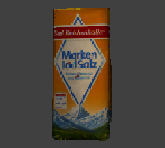}
  \hspace{5pt}
  \includegraphics[width=0.2\linewidth]{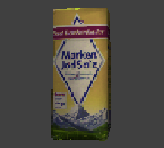}
  
  \vspace{5pt}
  
  \includegraphics[width=0.45\linewidth]{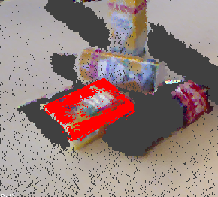}
  \hspace{5pt}
  \includegraphics[width=0.45\linewidth]{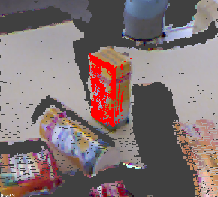}
  
  \caption{Pose estimation results from our setup. Top: input Kinect scene. Middle: input textured KIT objects (leftmost instances) and their color calibrated versions (rightmost instances). Bottom: estimation results for both objects, aligned objects overlaid in red.}
  \label{fig:realexp}
\end{figure}

In \reffig{fig:realexp}, we show pose estimation results for two objects. For completeness, we show both the original and the color calibrated version of an object, which is used during estimation. Note how a fairly large change in color representation is produced by the calibration routine. The estimated object pose is accurate enough for robotic manipulation via grasping.

%%%%%%%%%%%%%%%%%%%%%%%%%%%%%%%%%%%%%%%%%%%%%%%%%%%%%%%%%%%%%%%%%%%%%%%%%%%%%%%%

\section{Conclusions and future work}\label{conclusionsandfuturework}
We have presented a system for accurately estimating the alignment pose between two 3D models. Our model description is based on both appearance and shape data. Using an Early Cognitive Vision system, we split the input datum into separate representations in the edge and in the texture domains. By defining context descriptors on top of these visual modalities, we achieve a highly discriminative interpretation of a scene.

For utilizing our representation in an efficient manner, we have presented a RANSAC-based estimation procedure, which is based on fast rejection of outlier correspondences using low-level geometric constraints. Our experiments indicate that this makes the estimation process almost 15 times faster than standard RANSAC, without compromising estimation accuracy.

We have shown by quantitative evaluations that our descriptors perform almost twice as good as state of the art descriptors in the image domain and in the shape domain, thus making our context-based descriptors highly discriminative. For ten very different object classes, we achieve an average correspondence score of \unit{27}{\%}.

Finally, we have shown the practical usability of our system by accurate alignment results in two real setups. The system is currently being used in robotic grasping applications at our institute, and in the future we plan on using our system for accurate calibration between multiple cameras and robots.

In future works, a thorough investigation should be done in order to optimize the formulation of the context descriptor for ECV features. In this work, we have chosen a straightforward histogramming approach, incorporating simple angular and RGB relations. Although this has proven efficient, the context descriptor can undoubtedly be improved by alternative geometry- and appearance-based differential metrics, possibly using the local magnitude, orientation and phase which are right now used only for local image structure classification.

%%%%%%%%%%%%%%%%%%%%%%%%%%%%%%%%%%%%%%%%%%%%%%%%%%%%%%%%%%%%%%%%%%%%%%%%%%%%%%%%

\addtolength{\textheight}{-150mm}   % This command serves to balance the column lengths

%%%%%%%%%%%%%%%%%%%%%%%%%%%%%%%%%%%%%%%%%%%%%%%%%%%%%%%%%%%%%%%%%%%%%%%%%%%%%%%%

\section*{Acknowledgments}
This work has been supported by the EC project IntellAct (FP7-ICT-269959).

%%%%%%%%%%%%%%%%%%%%%%%%%%%%%%%%%%%%%%%%%%%%%%%%%%%%%%%%%%%%%%%%%%%%%%%%%%%%%%%%

\bibliographystyle{ieeetr}
\bibliography{ICRA2013_PoseEstimationLocalStructureSpecific}

%%%%%%%%%%%%%%%%%%%%%%%%%%%%%%%%%%%%%%%%%%%%%%%%%%%%%%%%%%%%%%%%%%%%%%%%%%%%%%%%

\end{document}